\documentclass{article}

\usepackage{arxiv}

\usepackage[utf8]{inputenc} % allow utf-8 input
\usepackage[T1]{fontenc}    % use 8-bit T1 fonts
\usepackage{hyperref}       % hyperlinks
\usepackage{url}            % simple URL typesetting
\usepackage{booktabs}       % professional-quality tables
\usepackage{amsfonts}       % blackboard math symbols
\usepackage{nicefrac}       % compact symbols for 1/2, etc.
\usepackage{microtype}      % microtypography
\usepackage{lipsum}
\usepackage[symbol]{footmisc}
\usepackage{graphicx}
\usepackage[english]{babel}
\usepackage[nottoc]{tocbibind}
\usepackage{amsmath}
\title{Scaling Traffic Insights with AI and Language Model–Powered Camera Systems for Data-Driven Transportation Decision Making}

\author{
  \textbf{Fan Zuo*, Ph.D.}\\
C2SMART Center,\\
Department of Civil and Urban Engineering\\
Tandon School of Engineering\\
New York University\\
\texttt{fz380@nyu.edu}\\
   \And
  \textbf{Donglin Zhou*, M.Sc}\\
 C2SMART Center,\\
Department of Civil and Urban Engineering\\
Tandon School of Engineering\\
New York University\\
  \texttt{dz2529@nyu.edu}\\
  \And
 \textbf{Jingqin Gao, Ph.D. (Corresponding author)}\\ %{\dag} \textsuperscript{\dag}\textit{Corresponding author.}
 C2SMART Center, \\
Department of Civil and Urban Engineering,\\
Tandon School of Engineering\\
New York University\\
  \texttt{jingqin.gao@nyu.edu}\\
  \AND
  \textbf{Kaan Ozbay, Ph.D.}\\
 C2SMART Center,\\
Department of Civil and Urban Engineering\\
Tandon School of Engineering\\
New York University\\
  \texttt{kaan.ozbay@nyu.edu}\\
\\
\\
\\
\\
\\
\\
\\
\\
\\
\\
\\
\\
\\
\\
\\
\\
\\
\\
}

\begin{document}

\maketitle

\thanks{*These authors contributed equally to this work.}
\newpage
\begin{abstract}
Accurate, scalable traffic monitoring is critical for real-time and long-term transportation management, particularly during disruptions such as natural disasters, large construction projects, or major policy changes like New York City (NYC)'s first-in-the-nation congestion pricing program. However, widespread sensor deployment remains limited due to high installation, maintenance, and data management costs. While traffic cameras offer a cost-effective alternative, existing video analytics struggle with dynamic camera viewpoints and massive data volumes from large camera networks. This study presents an end-to-end AI-based framework leveraging existing traffic camera infrastructure for high-resolution, longitudinal analysis at scale. A fine-tuned YOLOv11 model, trained on localized urban scenes, extracts multimodal traffic density and classification metrics in real time. To address inconsistencies from non-stationary pan–tilt–zoom (PTZ) cameras, we introduce a novel graph-based viewpoint normalization method. A domain-specific large language model (LLM)was also integrated to process massive data from a 24/7 video stream to generate frequent, automated summaries of evolving traffic patterns, a task far exceeding manual capabilities. We validated the system using over 9 million images from roughly 1,000 traffic cameras during the early rollout of NYC congestion pricing in 2025. Results show a 9\% decline in weekday passenger vehicle density within the Congestion Relief Zone, early truck volume reductions with signs of rebound, and consistent increases in pedestrian and cyclist activity at corridor and zonal scales. Experiments showed that example-based prompts improved LLM's numerical accuracy and reduced hallucinations. These findings demonstrate the framework’s potential as a practical, infrastructure-ready solution for large-scale, policy-relevant traffic monitoring with minimal human intervention. 
\end{abstract}

% keywords can be removed

\section{Introduction}
The availability of accurate, timely, and trackable traffic data is fundamental to effective transportation management \cite{FHWAtravelmonitoring}. These metrics often form a core component of advanced traffic management systems (ATMS), supporting real-time operations such as traffic incident management (TIM), adaptive signal control, and both short- and long-term planning. Recent advances in artificial intelligence (AI), particularly computer vision, have enabled automated extraction of traffic information from video and images, providing valuable information in monitoring localized areas or individual corridors with minimal human intervention.

Despite the growing maturity of AI-driven monitoring systems in localized contexts, scaling such approaches to a citywide or regional level introduces several critical challenges. Deploying a new AI-based, purpose-built video monitoring system from the ground up is financially and logistically burdensome for large urban areas. However, most cities already maintain an extensive intelligent transportation system (ITS) infrastructure of closed-circuit television (CCTV) cameras. Although these systems may have originally been deployed for manual monitoring, which relies on visual inspection or report-based inputs to support TIM and traveler information services, leveraging these existing assets offers a cost-effective path to large-scale AI-based monitoring. However, because legacy systems were not designed for automated data extraction, achieving agency objectives requires a robust post-processing pipeline.

One particular challenge is the prevalent use of pan-tilt-zoom (PTZ) cameras in existing ITS deployments. These cameras often do not maintain a fixed field of view, as they are routinely repositioned to serve diverse operational needs (e.g., emergency response). This variability causes inconsistent visual data, even for the same location, introducing substantial bias and errors in automated traffic metrics and making longitudinal trend analysis unreliable. Scalable AI-based monitoring frameworks must therefore account for this dynamic camera behavior to ensure consistent and valid data outputs.

The massive data generated by 24/7 traffic camera systems is another challenge. For example, a city such as New York uses about 1,000 traffic cameras in its daily operation. The volume of video data produced and information extracted can quickly exceed the capacity for manual review and interpretation by traffic engineers and planners. Automated tools and processing pipelines are needed to extract timely, interpretable insights on traffic patterns, vehicle classes, and performance trends without requiring exhaustive human supervision and validation.

To bridge these gaps, we developed a traffic monitoring tool that uniquely combines viewpoint normalization for highly variable PTZ cameras with deep learning-based detection and a large language model (LLM)-powered natural language summarization to support network performance evaluation and longitudinal analysis. We demonstrate its effectiveness through a use case in NYC during the initial months of congestion pricing implementation, a period when close and continuous monitoring was critically important. The main contributions of this study are:
\begin{itemize}
\item \textbf{Automated Traffic Monitoring Tool for Large-Scale Networks and Longitudinal Traffic Analysis}: We developed a deep learning–based tool that extracts traffic density and user classification data from real-time camera feeds, scalable to large networks. Validated on ~1,000 NYC cameras, it was used to study early impacts of the U.S.’s first congestion pricing program. This approach captures historically perishable data, filling a key gap by measuring within-zone traffic activities pre-and post-deployment, which supplements the managing agency's current evaluation method of counting vehicle entries.
    \item \textbf{Graph-Based Viewpoint Normalization for Dynamic Camera Feeds}: We introduced a novel viewpoint normalization technique to handle challenges posed by non-stationary PTZ cameras. This approach estimates camera tilt angles through homography and prioritizes dominant viewpoint groups using a graph-based clustering framework. In contrast to existing methods that rely on auto-recalibration or enforce fixed camera placements, our solution considers the operational needs of real-world surveillance systems, where cameras are prioritized to be reoriented to support emergency response efforts. By preserving data from stable and recurring viewpoints, the method enables consistent long-term monitoring and facilitates robust longitudinal traffic analysis.
    \item \textbf{LLM-Based Summarization with Domain-Specific Prompt Engineering}: We designed and tested an LLM module with different levels of structured prompting that transforms massive, heterogeneous video data into periodic, interpretable summaries of evolving traffic patterns (e.g., policy-driven route shifts), reducing the need for labor-intensive human interpretation.
    \item \textbf{Infrastructure-Compatible Design}: The proposed AI-based system is designed to work with existing CCTV infrastructure, eliminating the need for costly upgrades or the deployment of specialized AI camera systems.
\end{itemize}

\section{Literature Review}
\subsection{Recent Advances in Computer Vision for Transportation Object Detection}
Computer vision has transformed traffic monitoring by enabling tasks such as vehicle detection, classification, counting, and behavior analysis (e.g., risky behavior such as near misses) through state-of-the-art machine learning and deep learning models. Traditional approaches like deformable part-based models, such as histograms of oriented gradients (HOG) or local binary patterns (LBPs) \cite{nanni2017handcrafted} have largely been supplanted by convolutional neural network (CNN)–based object detection methods, such as the You Only Look Once (YOLO) family \cite{jiang2022review}, R‑CNN and Faster R‑CNN \cite{maity2021faster}, Single Shot MultiBox Detector (SSD) \cite{chen2022fast}, and RetinaNet \cite{hoang2019deep} due to their robustness and highly accurate end-to-end detection pipeline capabilities.

CNNs are designed to automatically learn and extract complex features from images during the training process to effectively interpret visual content. CNN-based architectures can typically be categorized into two types: two-stage models (e.g., R-CNN, Faster R-CNN) and single-stage models (e.g., YOLO, SSD, RetinaNet) \cite{khazukov2020real}. The former first generate region proposals and then classify those regions, generally achieving higher accuracy, while the latter perform detection and classification in a single step, which provides faster processing speeds at the cost of some precision. This trade-off between speed and accuracy often informs the choice of model depending on the specific goal of the application. For instance, Chen et al. proposed an occlusion-aware vehicle detection CNN model that incorporates local features and achieved a 3.17\% improvement over Faster R-CNN when tested on the Urban Traffic Surveillance Dataset (UTSD), which includes complex urban scenes and a high number of occluded vehicles.

Some studies have also explored hybrid architectures. For example, FCN‑rLSTM \cite{zhang2017fcn} combines fully convolutional networks with long short term memory networks (LSTM) layers to count vehicles over time using low-resolution city camera feeds, which improves density estimation accuracy while handling temporal dynamics. Building on these advances, more recent studies have increasingly incorporated transformer-based methods, for instance, Detection Transformer (DETR) \cite{carion2020end}, which originated in natural language processing but now widely applied to vision tasks like object detection. Unlike CNNs, which mainly capture local features, transformers tokenize inputs and use self-attention to model spatial and temporal dependencies, focusing on the most relevant information \cite{ghahremannezhad2023object}.

\subsection{Computer Vision-Based Traffic Monitoring System Using Existing Traffic CCTVs}
CCTV systems are among the most widely deployed ITS technologies for transportation management. According to the U.S. Department of Transportation (DOT)'s 2023 ITS Deployment Tracking Survey, 85\% of freeway agencies, 61\% of state DOTs, and 9\% of local arterial agencies use CCTV for incident detection, with many also relying on these feeds for monitoring traffic conditions and work zones \cite{petrella2023its}. A review of public camera networks found that over 60\% of U.S. city and state DOTs operate 50–500 cameras, and 7\% operate more than 1,000 \cite{gao2023exploring}. These systems are often well-funded; for example, Seattle’s CCTV network includes 286 cameras and costs over \$1 million annually to maintain \cite{seattle2021cctvaudit}.

Historically, video feeds have not been retained due to privacy, storage, and the lack of automated processing tools. However, recent advances in computer vision have spurred interest in extracting structured data from CCTV footage, with applications such as vehicle and pedestrian detection, signal control, and wrong-way driving alerts \cite{sarwar2024trafficlight, zuo2020interactive,zuo2021reference, lasky2020vision, haghighat2020wwd}. The U.S. DOT recognizes enhanced traffic CCTV analytics as a scalable, cost-effective data collection strategy for both urban and regional systems \cite{usdots_itsjpo_2022_datacollection}.

Despite advances in detection algorithms, the practical use of AI-based traffic camera analytics remains limited. Most research relies on small-scale pilots that are not fully operationalized. Some city and state agency projects \cite{martin2024raleigh, khalafi2024lasvegas} are still in early stages or limited in scope, while larger efforts like the Newcastle Urban Observatory \cite{newcastle2025urbanobservatory} and Chicago’s SAGE project \cite{sagecontinuum2025data} mainly serve researchers and developers rather than public-facing applications. A major technical hurdle is that most agency cameras are PTZ, which present challenges for automated video analytics due to shifting viewpoints \cite{ghahremannezhad2023object}. Prior studies have noted that many computer vision–based applications need fixed views or auto-calibration to ensure stable, reliable data extraction  \cite{ghahremannezhad2023object, haghighat2020wwd}.

In summary, while computer vision offers great potential for extracting actionable traffic insights from CCTV systems, key challenges remain, including treating non-stationary cameras, producing interpretable outputs, and enabling stable longitudinal analysis in real-world deployments.\\

\section{Methodology}
\subsection{Overview}
This paper proposes a modular and scalable system that integrates computer vision and LLMs to automate traffic monitoring from public traffic camera streams. The system consists of three components: (1) image collection and normalization; (2) automatic object detection and traffic density extraction; and (3) natural language summarization of traffic conditions from a very large dataset in real-time.

The system first collects and samples images from DOT cameras at fixed intervals in real-time. Because PTZ cameras are frequently reoriented, images from the same camera can vary greatly in viewpoint and zoom. To ensure consistency in subsequent analysis, we introduce a novel \textbf{viewpoint normalization} that combines scale-invariant feature transform (SIFT) and random sample consensus (RANSAC) to compute pairwise homographies of all images by each camera. These homography matrices are used to group images by viewpoint, allowing the system to isolate the largest stable subset of consistently oriented images for reliable temporal comparisons.

After viewpoint normalization, the system applies \textbf{deep learning-based object detection} to the images to extract traffic information on vehicles, pedestrians, and cyclists. The data is then converted into structured numerical metrics, including estimated densities and trends, for real-time or retrospective analysis.

Finally, an \textbf{LLM module} is used to generate automated textual summaries based on location, timestamp, and relevant metrics metadata to produce descriptive analysis. 

\begin{figure}[htbp]
  \centering
  \includegraphics[width=1.0\textwidth]{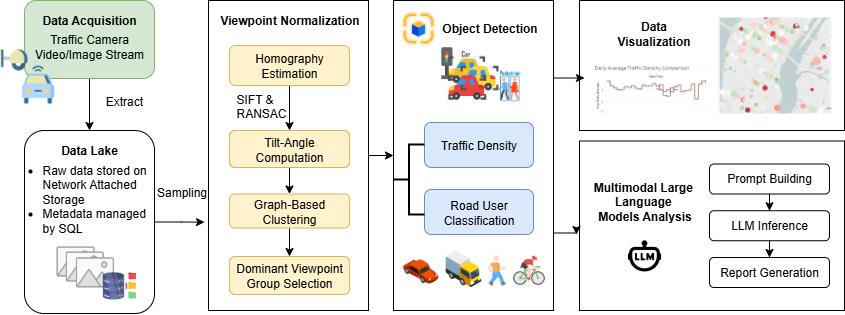}
  \caption{Overall architecture of the proposed methodology}
  \label{fig:flowchart}
\end{figure}

This architecture combines viewpoint normalization, deep learning-based detection, and LLM-driven reporting to transform raw public camera streams into structured, human-readable traffic insights at a citywide scale. It supports both local analyses, such as monitoring work zones or policy changes, and network-level evaluations of broader spillback effects. Figure~\ref{fig:flowchart} illustrates the overall architecture of the proposed methodology.

\subsection{Viewpoint Normalization via Homography Matching}
\begin{figure}[!ht]
  \centering
  \includegraphics[width=0.9\textwidth]{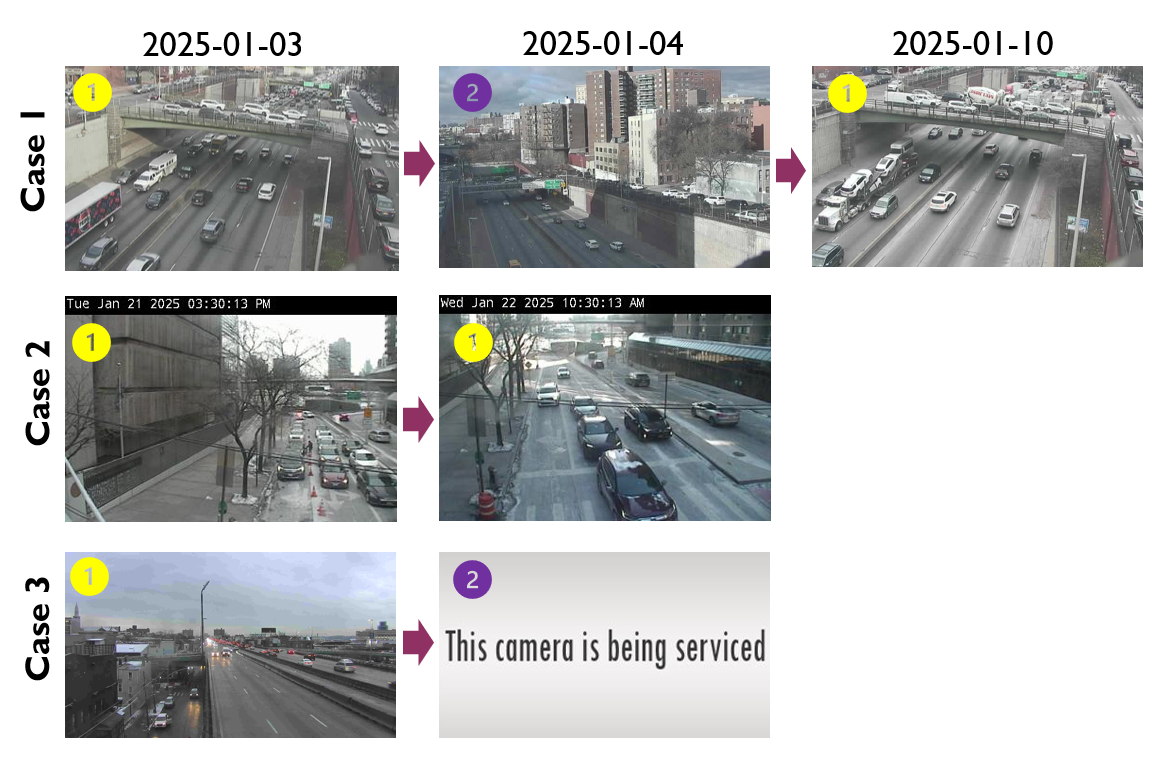}
  \caption{Illustrative examples of PTZ camera viewpoint variations: Case 1 – camera switches to a different roadway approach and later returns; Case 2 – same approach but with altered tilt or zoom, creating partial overlap with the original view; Case 3 – camera temporarily out of service.}\label{fig:viewpoint}
\end{figure}

PTZ cameras are very flexible in the field, yet that flexibility turns into a problem in the case of longitudinal computer‑vision analytics because images captured minutes apart can present entirely different roadway scenes; see Figure~\ref{fig:viewpoint} for three typical cases. Without correction, any "trend" we measure could be biased due to a tilt, a pan, or a zoom. To tame this nuisance, we introduce a \emph{viewpoint normalization} module that automatically groups frames with similar perspectives, allowing for a fair before-and-after comparison of traffic states.

\begin{figure}[!ht]
  \centering
  \includegraphics[width=1\textwidth]{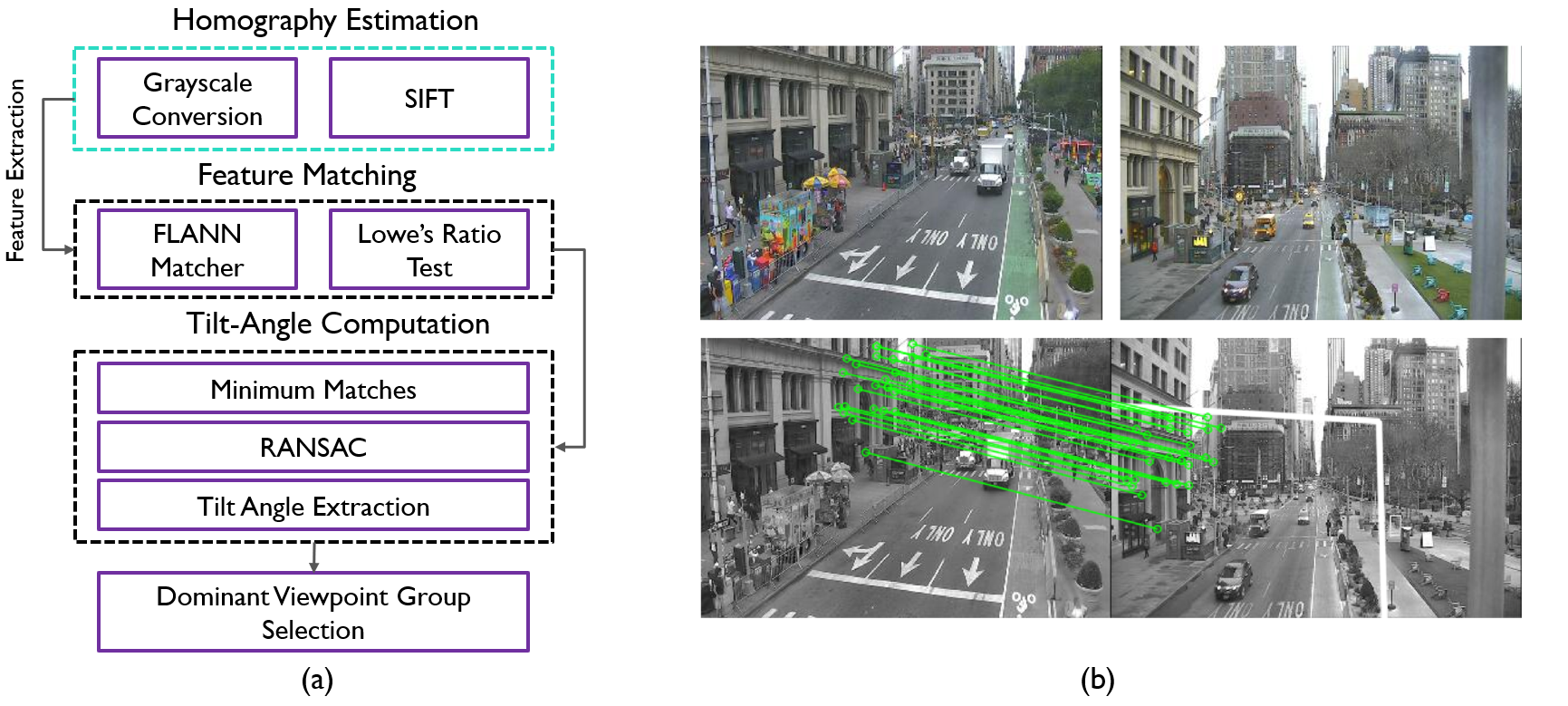}
  \caption{(a) Overview of the proposed viewpoint normalization framework. (b) Example SIFT matches (green lines).}\label{fig:keypointmatch}
\end{figure}

The normalization process is based on pairwise homography estimation, which quantifies the geometric transformation between two images. Two core computer vision techniques are employed: SIFT and RANSAC. SIFT is used to extract local keypoints in each image, which are invariant to rotation, scale, and minor affine transformations. RANSAC is then applied to match these keypoints across image pairs and to compute a homography matrix by eliminating outlier correspondences. Figure~\ref{fig:keypointmatch} presents the overall workflow of the proposed viewpoint normalization method along with an example of keypoint matching.

\noindent\textbf{Algorithmic details.}  
Let $I_i, I_j$ denote two frames from the same camera. A homography $\mathbf{H}_{ij}\!\in\!\mathbb{R}^{3\times3}$ satisfies
$\mathbf{x}'\!\sim\!\mathbf{H}_{ij}\mathbf{x}$ for corresponding homogeneous points
$\mathbf{x},\mathbf{x}'$. We perform:

\begin{enumerate}%[wide, labelwidth=!, labelindent=12pt, leftmargin=3em]
  \item \emph{Matching \& Homography fit}: detect $\approx\!2000$ SIFT keypoints per frame; keep matches passing Lowe’s test; fit $\mathbf{H}_{ij}$ with RANSAC (1\,000 iterations, inlier threshold $2$\,px).  We retain the solution if the inlier ratio exceeds $0.25$, a value calibrated on a dev set of 10000 frame pairs.
  \item[•] \emph{Tilt‑angle extraction}:  
        Write $\mathbf{H}_{ij}=\bigl[\mathbf{R}\ \mathbf{t}\bigr]$ with
        $\mathbf{R}\in\mathbb{R}^{2\times2}$ the affine approximation.
        After scale removal
        \begin{equation}
            \mathbf{R}_{\text{norm}}
              = \frac{\mathbf{R}}{\sqrt{h_{11}^2+h_{21}^2}},
              \qquad
              \theta_{ij}
              = \operatorname{atan2}(r_{21},r_{11})\cdot\frac{180}{\pi}, 
        \end{equation}
        we tag $(i,j)$ as "same view" if $|\theta_{ij}|\le\delta$ with
        $\delta = 10^{\circ}$.  
        Empirically, this threshold balances false merges ($<4\%$) and false splits ($<6\%$).
  \item[•] \emph{Graph clustering}:  
        Build an undirected graph $G=(V,E)$ where $V$ is the frame set and
        $(i,j)\!\in\!E$ whenever $|\theta_{ij}|\le\delta$.  
        Connected components are the viewpoint clusters
        $\{\mathcal{C}_k\}$.
  \item[•] \emph{Dominant‑view selection}:  
        Let $k^\star=\arg\max_k|\mathcal{C}_k|$. Only frames in
        $\mathcal{C}_{k^\star}$, the camera’s predominant view, are forwarded to the detection pipeline; the remainder are discarded.
\end{enumerate}
\hfill\break%
This approach ensures that only images taken from stable and consistent perspectives are used in traffic density estimation, where highly similar views are clustered despite being recorded at different times. Once grouped, these normalized images are forwarded to the detection pipeline for traffic object extraction. 

\subsection{Object Detection and Automated Data Pipeline}
Once every frame has been parked in its “stable‑view” bucket, it flows into a fully automated detection pipeline that converts pixels to numbers we can actually crunch. The goal is to translate raw CCTV imagery into machine-readable traffic metrics at the city scale without human intervention.

\begin{enumerate}%[wide, labelwidth=!, labelindent=12pt, leftmargin=3em]
    \item[•] \textbf{Object detection \& data extraction.} We deploy a \textbf{YOLOv11‐L} network (trainable parameters
          $P\!\approx\!43$M) fine‑tuned for urban traffic scenes.  
          Given an input frame $I_t$, the detector outputs a set
          $\mathcal{B}_t = \{(b_k, c_k, s_k)\}_{k=1}^{N_t}$ of bounding boxes
          $b_k\in\mathbb{R}^4$, class labels $c_k\in\{\,\text{car},\text{truck},\text{ped},\text{bike}\,\}$, and confidence scores $s_k\in[0,1]$.  
          We keep detections with $s_k\ge0.35$, an empirically chosen operating point that maximizes the F\textsubscript{0.5} measure on our validation set.  
          Aggregate counts per class are
          \begin{equation}
               n_{t}^{(c)} = \sum_{k=1}^{N_t}\mathbf{1}\{c_k = c\},
            \qquad c\in\mathcal{C},
          \end{equation}
          and frame‑level density (objects/$\text{m}^2$) is estimated via
          \(
            d_t^{(c)} = \frac{n_t^{(c)}}{A_{\text{ROI}}}\!,
          \)
          where $A_{\text{ROI}}$ is the calibrated road‑surface area in
          $\text{m}^2$.
    \item[•] \textbf{Data aggregation.}  
          Each detection packet is flattened into a vector: [$camID$, $t$, $n_t^{(\text{car})}$, $n_t^{(\text{truck})}$, $n_t^{(\text{ped})}$, $n_t^{(\text{bike})}$, $vpID$] 
          and streamed to a PostgreSQL database.
    \item[•] \textbf{Interactive visualization.}  
          We plug the database into a Tableau workbook dubbed \emph{Multimodal Density Tracker}. Key widgets include: (i) a time‑slider interactive map with colored circles represent average weekly traffic density and its changes for each camera location; (ii) bar charts showing daily average density by day of week for 2024 and 2025; (iii) data type dropdown that allows switching between road user types.\\
\end{enumerate}

Object detection is performed using an enhanced YOLOv11s model with a two-stage training process using the VisDrone dataset \cite{zhu2021detection} and a localized dataset generated by the research team of manually labeled New York City (NYC) street scenes, referred to as UTO (Urban Transportation Objects). The VisDrone dataset includes 288 video clips containing 261,908 frames, along with 10,209 static images captured by drone-mounted cameras in various urban environments \cite{zhu2021detection}. It is particularly well-suited for detecting small objects due to its aerial perspectives and high object density. The localized dataset includes 5,456 traffic camera images with 60,494 annotated objects (people, bicycles, cars, trucks, and buses), and was developed to improve detection accuracy and reduce false positives in the context of a complex urban environment. The model was first pretrained on the VisDrone 2019 dataset and subsequently fine-tuned on a custom-labeled dataset. To preserve low-level feature representations learned from the VisDrone dataset, we froze the first five layers of the YOLOv11s backbone during fine-tuning. The remaining layers were trained on our custom dataset to adapt the model to the specific camera views and object scales present in the CCTV cameras.

\subsection{LLM-based Traffic Summarization}
\begin{figure}[!ht]
  \centering
  \includegraphics[width=1.0\textwidth]{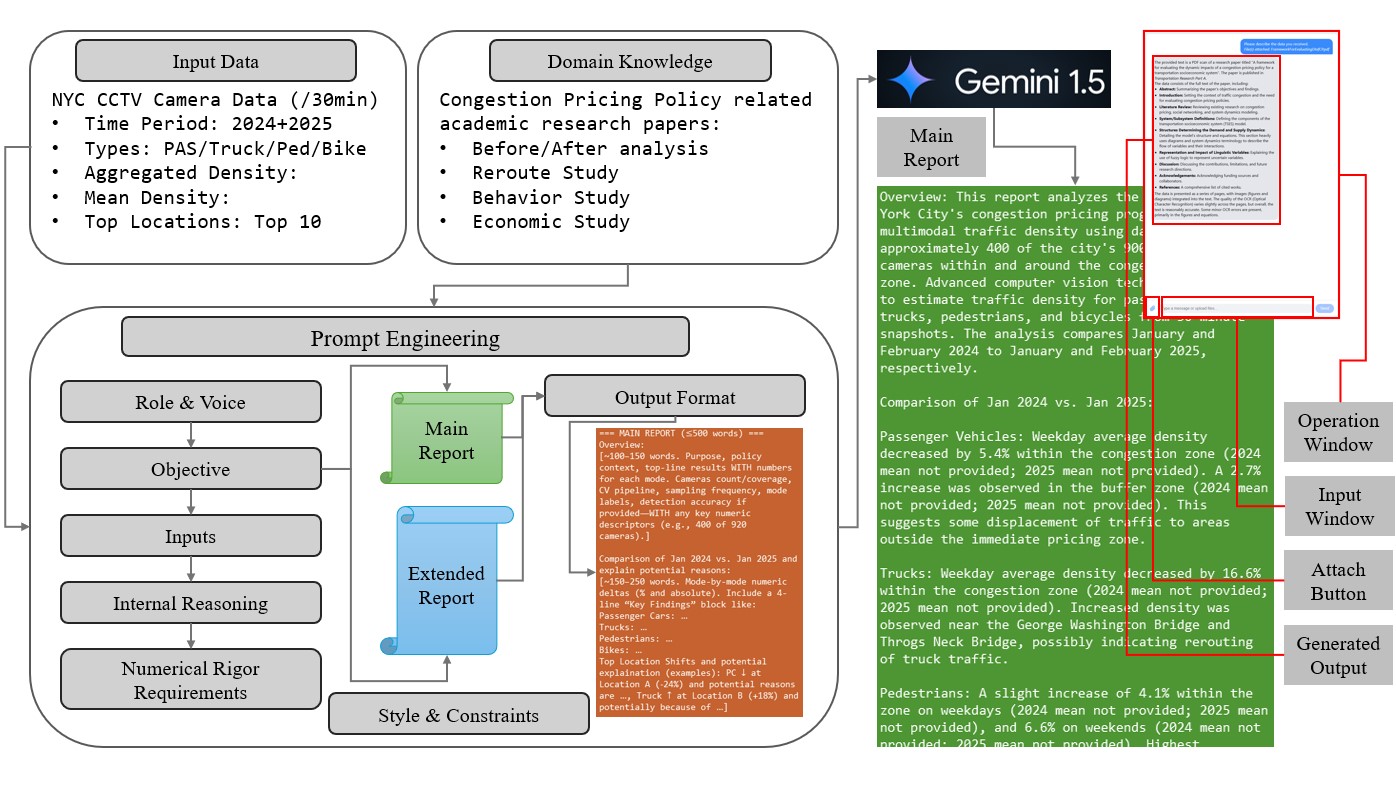}
  \caption{LLM‑Augmented Traffic Summarization Workflow.}
  \label{fig:llm_workflow}
\end{figure}

Camera-derived data must be analyzed to understand changes in traffic densities; however, manually synthesizing information from 1,000 continuously operating cameras is infeasible for human analysts. Moreover, answering targeted decision-making questions, for example, identifying potential route shifts for truck traffic, becomes even more labor-intensive. To translate structured traffic metrics into human-readable summaries and analysis, we integrate LLMs to generate summaries of observed traffic patterns and diagnostic reasoning about the underlying causes. This section details our prompt-engineering pipeline and evaluation protocol for generating interpretable, policy-facing summaries from large-scale CCTV-derived multimodal traffic data. We target two complementary outputs: (i) a succinct, general-audience brief (\(\leq\)500 words) with mandatory modal and spatial statistics; and (ii) an extended technical report with full numerical details. The pipeline is designed to be reproducible with data insights. We standardize on a single backbone model, \textit{Gemini~1.5}, and evaluate how different levels of structured \emph{prompting}, from simple requests to domain-informed instructions, with or without numerical specifications, affects model accuracy and reliability. Figure \ref{fig:llm_workflow} shows the system pipeline using evaluation of a before-and-after policy deployment as an example.
\subsubsection{Input Preparation and Prompt Design}

For each timestamp–camera record, we include mode-specific densities for various road user classes. The raw dataset includes camera identifiers, geocoordinates, borough, and zone flags (\textit{inside}, \textit{boundary}, \textit{outside}). Prior to LLM ingestion, we perform:\\
\begin{enumerate}%[wide, labelwidth=!, labelindent=12pt, leftmargin=3em]
    \item[•] \textbf{Temporal harmonization:} aligning analysis windows between pre- and post-deployment, matching weeks, weekdays/weekends, and peak/off-peak periods. Missing days are handled via listwise deletion for headline comparisons and via sensitivity checks.

    \item[•] \textbf{Aggregation schemas:} (a) per-camera statistics; (b) zone aggregates; (c) borough aggregates; (d) "top changes" lists: top-\(K\) locations with increases and decreases per mode with absolute (\(\Delta\)) and relative (\%\(\Delta\)) changes.

    \item[•] \textbf{Summary features:} for each mode \(m\in\{\text{car, truck, ped, bike}\}\) and partition \(p\) (zone/camera): totals, mean, median, and standard deviation.\\
\end{enumerate}

We evaluate four prompt configurations that progressively strengthen structure, numerics, and domain grounding. All configurations request two outputs: a \emph{Main Report} (\(\leq\)500 words) with sections \textit{Overview}, \textit{Description of Data}, and \textit{Comparison of 2024 vs.\ 2025}; and an \emph{Extended Report} with unconstrained length. The four configurations are:\\
\begin{enumerate}%[wide, labelwidth=!, labelindent=-5pt]
    \item[•] \textbf{Stage~A (baseline): No prompt engineering (simple prompt).} A short, generic instruction to ``summarize trends'' from the provided CSV statistics. No explicit section headings, no numeric requirements, and no domain cues.
    \item[•] \textbf{Stage~B: Prompt engineering (structure only).} We require the two reports to follow the \emph{exact} section headers, a general-audience tone, and explicit coverage of all four modes and spatial highlights. However, we do not enforce the use of quantitative formulas or domain-specific exemplars.
    \item[•] \textbf{Stage~C: Prompt engineering with detailed quantitative instructions (no training on domain knowledge).} We add strict guidelines: every major claim must include 2024 and 2025 numeric values, absolute change, and percentage change with units; we also require peak, weekday/weekend splits where available, and top-\(K\) location lists with highest and lowest values. No domain exemplars are used.
    \item[•] \textbf{Stage~D: Prompt engineering with detailed quantitative instructions \emph{and} domain exemplars.} Building on Stage~C, we augment the prompt with short expert-written mini-briefs and retrieval-augmented snippets distilled from peer-reviewed congestion pricing studies (e.g., spillovers, rerouting, temporal heterogeneity, industry effects). Retrieved text is capped to top-\(k\) chunks per theme to avoid prompt bloat.
\end{enumerate}
\hfill\break
To improve determinism and reduce hallucinations, we set temperature \(\in[0.0,0.3]\), top-\(p\in[0.8,1.0]\), and request \(n\)-best candidates (\(n=2\)–3) for tie-breaking. We validate numeric strings against precomputed statistics before accepting an output. If a value is out of tolerance, the run is rejected and automatically re-prompted with corrective hints (e.g., "Your trucks \%\(\Delta\) exceeds the allowed range; recompute using the provided totals").

\subsubsection{Test and Performance Evaluation}
Once the input prompt is constructed, it is passed to the LLM for inference. We assess Gemini~1.5 under the four prompt configurations using three complementary metrics focused on numerical fidelity, content coverage, and safety:\\

\noindent\textbf{Numeric Consistency Score (NCS):} For each reported scalar \(y\) and ground-truth value \(g\), we compute the relative error
\begin{equation}
    \varepsilon = \frac{|y-g|}{\max(1,|g|)}.
\end{equation}
Let \(\mathcal{Y}\) be the set of required numeric items (totals, means, peaks, \(\Delta\), \%\(\Delta\), etc.). We define:
\begin{equation}
    \mathrm{NCS} = 1 - \frac{1}{|\mathcal{Y}|}\sum_{(y,g)\in\mathcal{Y}} \min(\varepsilon, 1.0),
\end{equation}
so that \(\mathrm{NCS}\in[0,1]\), where higher values indicate better agreement.\\
\noindent\textbf{Content Match F1 (CM-F1):} We construct an expert ground-truth (GT) checklist of \(K\) key findings that span modes, spatial hotspots, and temporal patterns. Let \(P\) be the set of model-reported findings. Using fuzzy semantic matching with numeric tolerance \(\tau\) (e.g., \(\pm 1\) percentage point for \%\(\Delta\)), we compute
\begin{equation}
    \mathrm{Precision}=\frac{|P\cap_{\tau}\mathrm{GT}|}{|P|},\quad
    \mathrm{Recall}=\frac{|P\cap_{\tau}\mathrm{GT}|}{|\mathrm{GT}|},\quad
    \mathrm{CM\text{-}F1}=\frac{2\,\mathrm{Precision}\cdot \mathrm{Recall}}{\mathrm{Precision}+\mathrm{Recall}}.
\end{equation}
\noindent\textbf{Hallucination Rate (HR):} We flag any numeric or spatial assertion that lacks support in the provided statistics or violates tolerance thresholds:
\begin{equation}
    \mathrm{HR}=\frac{\text{\# unsupported claims}}{\text{\# total claims}}.
\end{equation}
Lower is better.\\

A weighted score function is used to evaluate the overall performance of the proposed model:\\
\begin{equation}
    \mathrm{Score_{model}} = 
    0.40\,\mathrm{NCS}+
    0.40\,\mathrm{CM\text{-}F1}+
    0.20\,(1-\mathrm{HR})
\end{equation}\\

\section{Case Study: Evaluating Early Impacts of NYC's Congestion Pricing Program}
To demonstrate the real-world applicability of the proposed methodology and validate its practical value, we applied it to approximately 1,000 publicly available traffic cameras in NYC to analyze changes in traffic patterns before and after the launch of the congestion pricing program.

\subsection{Overview of NYC's Congestion Pricing Program}
Traffic congestion arises when demand exceeds roadway capacity, a persistent problem in large cities driven by population growth, limited infrastructure, and rising vehicle ownership. In 2024, traffic congestion cost the U.S. economy over \$74 billion, with NYC ranking first nationwide and second globally for delays—drivers lost 102 hours annually, incurring an average cost of \$1,826 \cite{inrix2024}.

\begin{figure}[!ht]
  \centering
  \includegraphics[width=0.8\textwidth]{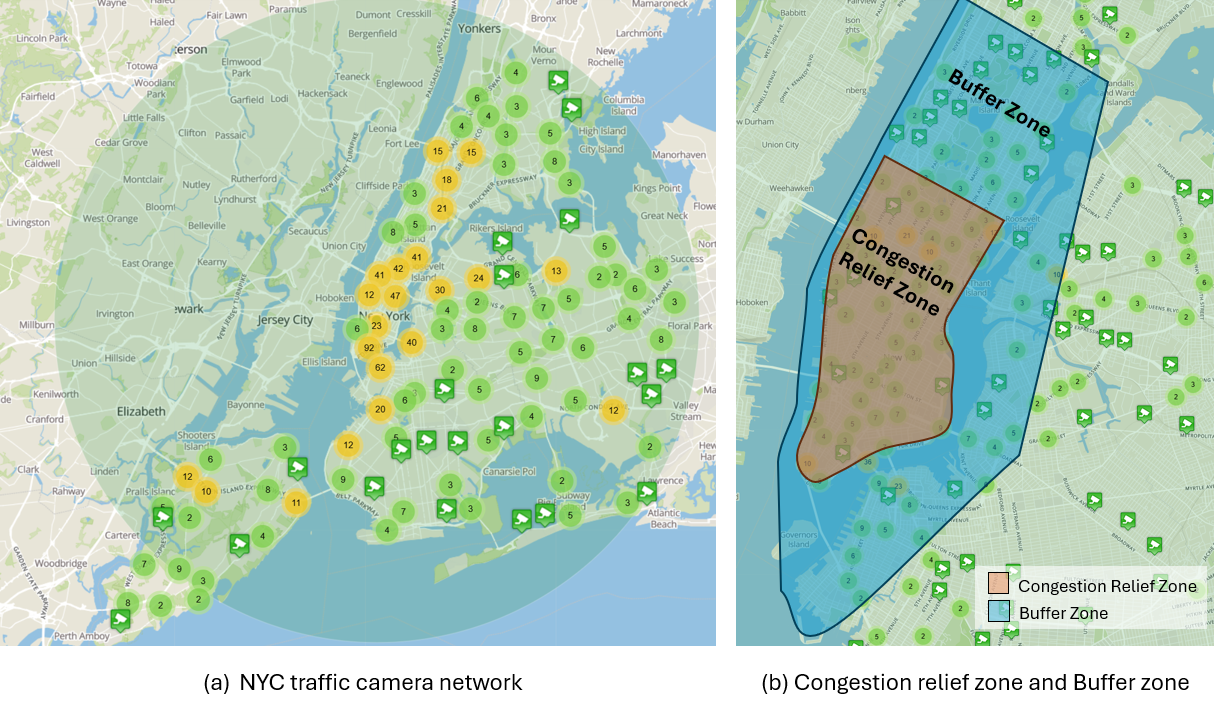}
  \caption{NYC traffic camera network and congestion pricing toll zone}\label{fig:cameranetwork}
\end{figure}

Congestion pricing is a market-based strategy that charges drivers to enter high-demand zones during peak hours, encouraging rerouting, rescheduling, or mode shifts \cite{fhwaCongestionPricing}. On January 5, 2025, the Metropolitan Transportation Authority (MTA) implemented the first U.S. program of this kind, charging vehicles to enter the Congestion Relief Zone (CRZ)—Manhattan streets at or below 60th Street (Figure~\ref{fig:cameranetwork} (b)) \cite{mta_cbdtp_2025}. Rates vary by vehicle type, time, payment method, and crossing credits. Generally, most passenger cars pay \$9 during peak hours, heavy vehicles \$14.40–\$21.60 (with 75\% overnight discounts), and taxis and for-hire vehicles \$0.75 and \$1.50 per trip, respectively \cite{mta_cbdtp_2025}.

Although the MTA has deployed 110 camera-based sensors to record vehicle entries into the CRZ, these systems primarily track entry counts \cite{mta_cbdtp_2025}. Comprehensive data on traffic volume and density across the city remains sparse, since existing sources are typically derived from one-time data collection efforts that lack the temporal resolution necessary to capture day-to-day variations. Furthermore, baseline volume and density data prior to 2025 are limited, making the 2024 dataset produced by the proposed computer vision–based method a valuable and novel contribution to the empirical study of urban traffic dynamics. This case study aims to answer the following research questions:
\begin{itemize}
    \item[•] How did traffic patterns by mode and vehicle class change within the CRZ and nearby areas after congestion pricing began?
    \item[•] Are there measurable differences at major entry crossings such as bridges and tunnels?
    \item[•] Do traffic patterns shift at non-CRZ crossings, particularly among freight traffic for which mode substitution or time-of-day adjustments are less practical?
\end{itemize}

\subsection{Experiment Design}
We collected live image snapshots from 936 public CCTV cameras \cite{nyctmc_webcams}, including 202 cameras inside the CRZ and 176 cameras in its surrounding area (buffer zone). These cameras are spread across different streets, providing granular coverage of the city's roadways (Figure~\ref{fig:cameranetwork} (a). 

Camera feeds provide 240p snapshots every 2–7 seconds. To facilitate a scalable, long-term analysis of traffic conditions while maintaining computational feasibility, we used traffic density, the average number of visible road users per snapshot, as the primary metric. For consistency and efficiency, density is computed using snapshots sampled at 30-minute intervals and aggregated over time windows (e.g., weekly averages) by road user category.

To evaluate the early effects of congestion pricing on traffic dynamics, we collected approximately eight months of CCTV imagery data covering January through April in both 2024 (pre-deployment) and 2025 (post-deployment). This totaled 9,120,384 images, with a small number of days in March unavailable due to intermittent server disruptions. Multimodal traffic density across four modes, namely passenger vehicles, trucks, pedestrians, and bicycles, was analyzed to investigate changes in density across different modes and spatial contexts.

All object detection training and inference procedures were conducted on a workstation equipped with AMD Ryzen 9 7950X 16-Core Processor and two NVIDIA A6000 Ada GPUs (48 GB VRAM each) to support efficient multi-GPU training and large-batch evaluation. The algorithms were implemented in Python using PyTorch, with OpenCV employed for image processing tasks. Viewpoint normalization was applied to all cameras to ensure that a dominant, consistent viewpoint was selected for each camera to enable meaningful comparative analysis. As part of this analysis, we also assessed camera viewpoint stability, defined as the proportion of image snapshots in which the camera maintains a consistent viewpoint relative to a predefined stable view. For example, a stability score of 80\% indicates that the camera retains a stable, expected orientation 80\% of the time, while during the remaining 20\% it may be facing a different direction. This measure helps quantify the extent to which each camera provides a reliable, fixed perspective over time, which is critical for ensuring consistency in comparative analyses.

For LLM analysis, we run Gemini~1.5 across Stages~A--D with identical precomputed statistics and analysis windows. For Stage~D we enable retrieval-augmented context from a curated library of congestion pricing studies; chunks are selected via cosine similarity to queries that target \emph{mode shifts}, \emph{zone spillovers}, \emph{temporal heterogeneity}, and \emph{industry impacts}. We perform a \(3\times\) self-consistency sweep (temperatures \(0.2/0.25/0.3\)) and select the top candidate per stage by (i) highest NCS, (ii) highest CM-F1, and (iii) lowest HR. 

\subsection{Result and Discussion}
\subsubsection{Camera Viewpoint Stability Results}
\begin{figure}[!ht]
  \centering
  \includegraphics[width=0.5\textwidth]{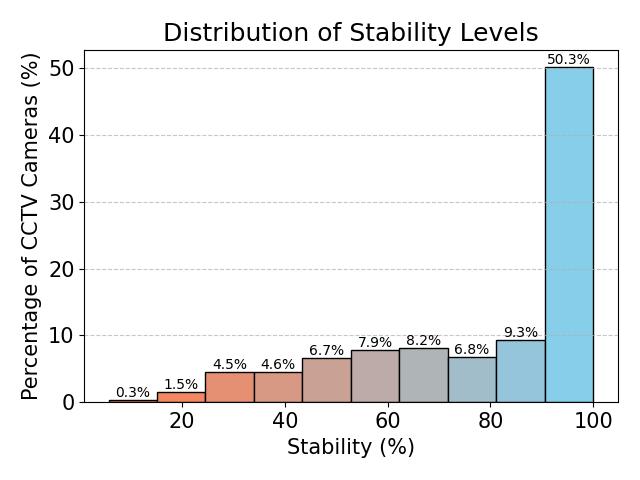}
  \caption{Camera viewpoint stability test results}\label{fig:stability}
\end{figure}

The camera viewpoint stability results show that more than half of the cameras maintain a fixed viewpoint over time, providing a strong foundation for consistent computer vision analysis (Figure~\ref{fig:stability}). The remaining cameras exhibit a gradual distribution across lower stability levels, ranging from 10\% to 90\%, which indicates that a subset of cameras experience occasional or frequent viewpoint shifts. Very few cameras fall below 20\% stability (i.e., complete instability), meaning that frequent view changes are rare in this dataset. In terms of spatial context, approximately 60\% of the cameras are oriented toward urban roadways and intersections, 32\% (252 cameras) monitor highways, and 8\% (56 cameras) are positioned at tunnels and bridges. This distribution reflects the diversity of the existing CCTV infrastructure and provides strong support for the applicability of the proposed approach for longitudinal and location-specific traffic analysis.

\subsubsection{Object Detection Results}
\begin{figure}[htbp]
  \centering
  \includegraphics[width=1.0\textwidth]{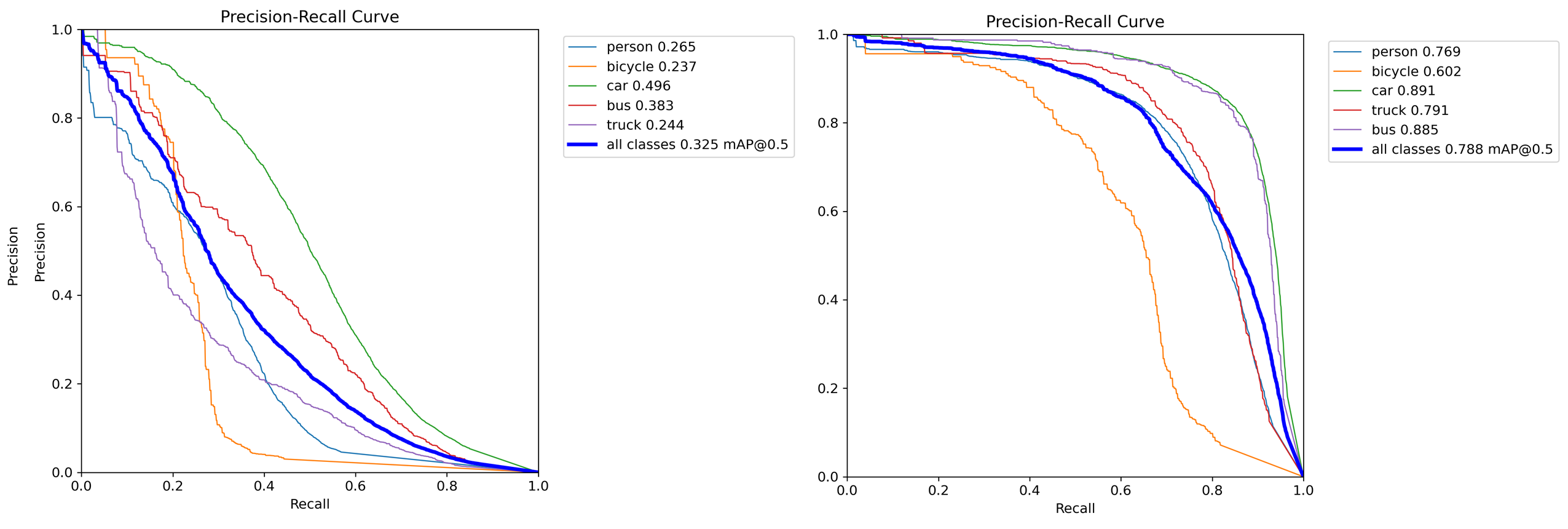}
  \caption{Model performance for (a) original YOLO v11s trained by the VisDrone dataset, and (b) enhanced YOLO v11s trained by the VisDrone + UTO}
  \label{fig:mAP}
\end{figure}

\begin{figure}[htbp]
  \centering
  \includegraphics[width=1.0\textwidth]{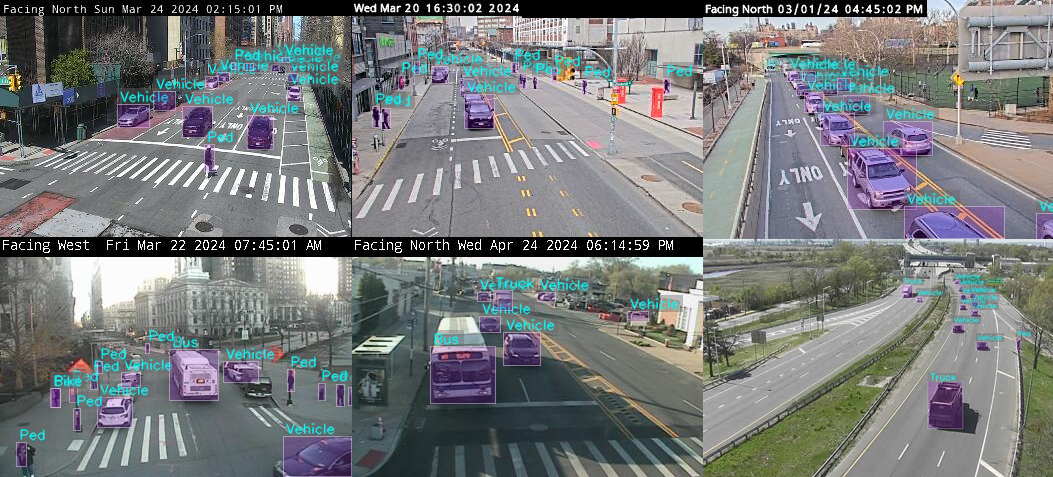}
  \caption{Examples of detection outputs}
  \label{fig:detectionoutput}
\end{figure}

Figure~\ref{fig:mAP} shows the mean average precision (mAP) at 0.5 for each road user class, comparing the original YOLOv11s model trained on the VisDrone dataset with the enhanced YOLOv11s trained on VisDrone and UTO. The object detection model achieved a mAP@0.5 of 0.788 across the classes of interest. The addition of training using a localized dataset UTO appears to improve model performance significantly across all classes, with notable gains in underperforming categories like bicycle and truck. UTO helped the model generalize or learn road user-specific features better, likely due to better representation in the enhanced dataset. Figure~\ref{fig:detectionoutput} shows visual examples of the detection outputs.

\subsubsection{LLM Analysis Results}
The LLM’s Main Report delivers a concise, 500‑word brief with the prescribed \emph{Overview}, \emph{Description of New Data}, and \emph{Comparison of 2024 vs. 2025} sections, each packed with exact 2024/2025 values, absolute deltas, percentage changes, and top‑K location highlights for every traffic mode. The Extended Report unfolds into a detailed, no‑limit narrative that systematically covers the background, data, and methodology, numeric results disaggregated by mode, time, and spatial zone, as well as a thorough discussion of policy implications. 
\begin{figure}[h]
  \centering
  \includegraphics[width=1.0\textwidth]{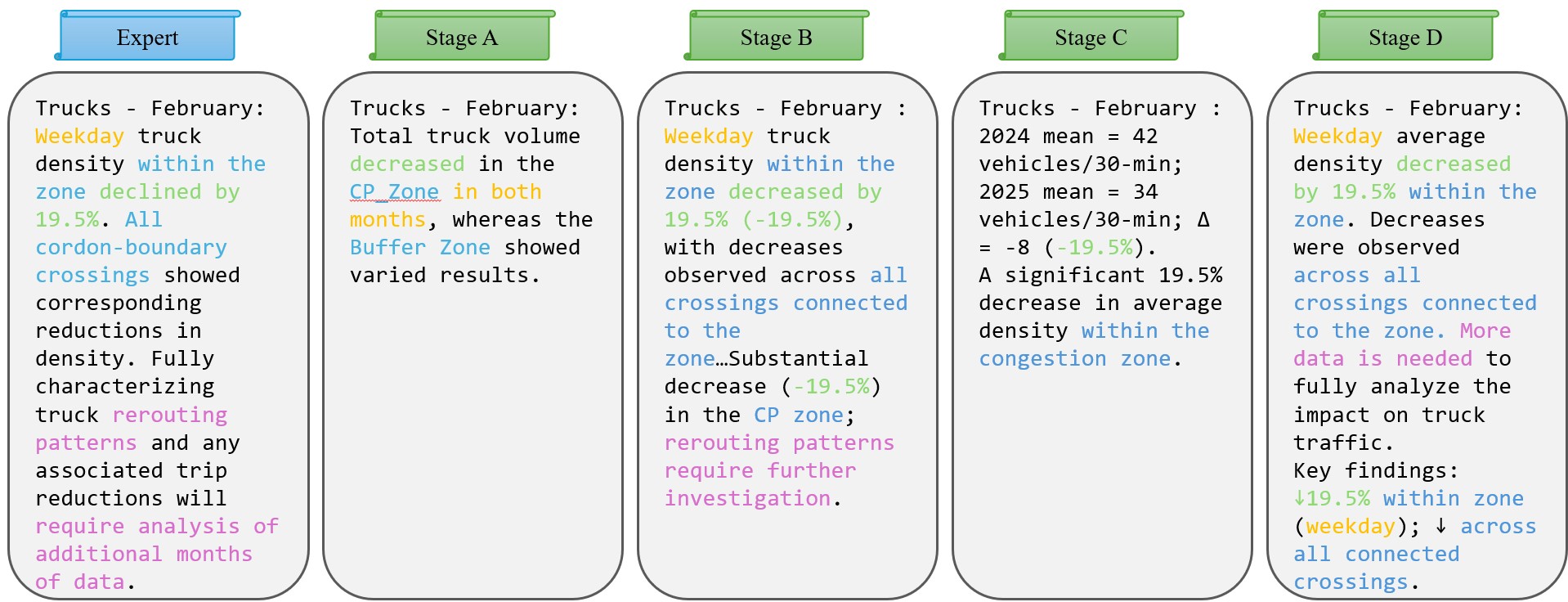}
  \caption{Comparison of Expert‑Generated and Prompt‑Stage Reports for February Truck Data Analysis}
  \label{fig:llm_comparison}
\end{figure}
Figure~\ref{fig:llm_comparison} illustrates how each prompt stage approximates the expert‑crafted February truck analysis by highlighting four information types: temporal references (yellow), spatial context (blue), numeric findings (green), and causal interpretation (purple). The expert report neatly integrates all four elements—precise time windows, affected locations, exact percentage changes, and plausible rerouting explanations—whereas Stage A barely touches on time or location and omits hard numbers and reasons entirely. Stage B recovers the section headings and mentions camera sites, but still lacks quantitative precision and interpretation. Stage C nails the green layer by reporting $\Delta$ and $\%\Delta$ but stalls on time specificity and offers no causal insight. Finally, Stage D comes closest to the expert: it aligns on time frames, pinpoints key bridges, delivers accurate statistics, and even conjectures about truck rerouting, though its reasoning is slightly more generic. This progression underscores that only the combination of strict numeric prompts and domain exemplars can faithfully reproduce the expert’s full analytical narrative.  
\begin{figure}[t]
    \centering
    \includegraphics[width=0.6\textwidth]{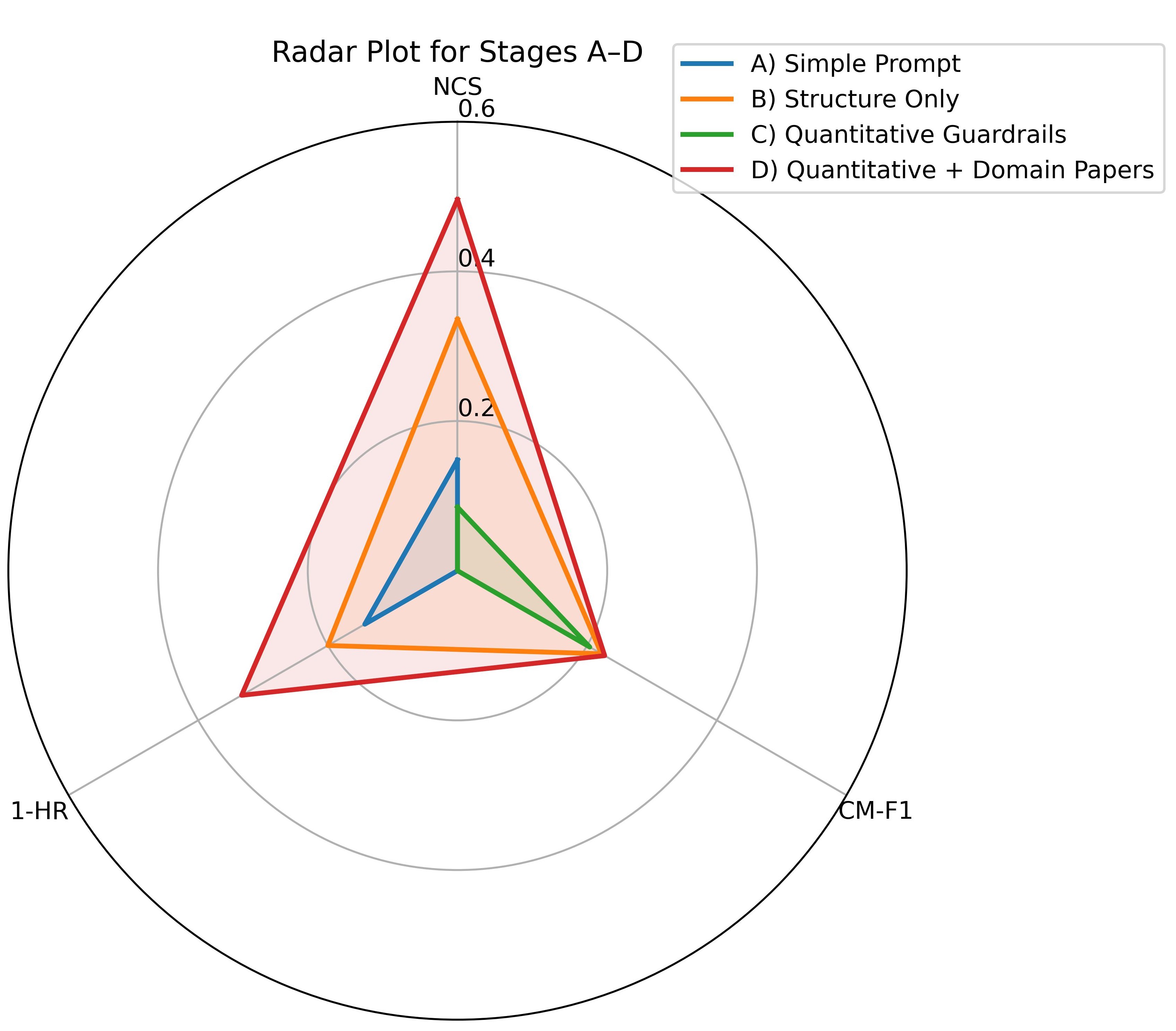}
    \caption{LLM performance under the four prompt configurations. Higher is better for NCS and CM-F1; \(1-\)HR is plotted for consistency.}
    \label{fig:radar_stages}
\end{figure}
Figure~\ref{fig:radar_stages} provides a radar comparison of NCS, CM-F1, and \(1-\)HR across Stages~A--D, and Table~\ref{tab:stage_scores} lists the corresponding scores. In brief, we observed a clear monotonic improvement as prompts became more structured and data-aware. 
\begin{enumerate}%[wide, labelwidth=!, labelindent=-5pt]
    \item[•] \textbf{Stage~A} (simple prompt): With no structural or numeric guidance, the model provided fluent prose but \emph{almost no verifiable numbers} (NCS~0.15) and invented values freely (HR~0.86). Qualitative claims rarely matched the ground truth, which is reflected by the CM–F1 score (around 0).
    \item[•] \textbf{Stage~B} (structure only): Requiring the three section headings and explicit modal coverage \emph{doubled} numeric fidelity (NCS~0.34) and increased CM–F1 to 0.22, but hallucinations remain common (HR~0.80) because numbers are still optional.
    \item[•] \textbf{Stage~C} (quantitative instructions): Paradoxically, simply adding “give 2024, 2025, $\Delta$, $\%\Delta$” without domain examples \emph{hurt} NCS and drove HR to 1.0. Manual inspection revealed that Gemini is attempting to fabricate the full quartet for every mode, but mis–parsing the statistics bundle; every fabricated number failed in the tolerance check.  This highlights that numeric guidelines \emph{alone} are brittle: the model needs to see \emph{how} to weave supplied numbers into prose.
    \item[•] \textbf{Stage~D} (quantitative instructions + domain exemplars): Injecting two expert mini‑briefs plus targeted research paper snippets largely improved the performance. NCS tripled relative to Stage B and quintupled relative to Stage C as the model now replicated rather than re‑computed statistics.  Hallucinations reduced to two‑thirds of Stage A (HR~0.67). The composite score rose to 0.356, more than quadrupling the naïve baseline. Gains came mostly from \emph{correct numbers}; CM–F1 nudged up only slightly because the short general-audience briefs already covered most ground‑truth claims.
\end{enumerate}

\begin{table}[t]
    \centering
    \caption{Scores by prompt configuration.}
    \label{tab:stage_scores}
    \begin{tabular}{|l|c|c|c|c|}
        \hline
        Stage & NCS & CM-F1 & HR (\%) & Final Score\\
        \hline
        A) Simple Prompt & 0.148 & 0.000 & 0.857 & 0.088 \\
        B) Structure Only & 0.336 & 0.222 & 0.800 & 0.263 \\
        C) Quantitative Instructions & 0.085 & 0.204 & 1.000 & 0.116\\
        D) Quantitative + Domain exemplars & \textbf{0.496} & \textbf{0.227} & \textbf{0.667} & \textbf{0.356}\\
        \hline
    \end{tabular}
\end{table}

The results revealed that small, domain-specific examples combined with explicitly defined numeric expectations outperform general prompts that merely encourage quantitative responses. Basically, \emph{training high-quality examples and enforcing precise numerical results yields better results}. As shown in Figure~\ref{fig:radar_stages}, Stage D clearly outperforms other stages, particularly along the NCS dimension.

Across all stages, we observed two recurring failure modes.  
First, \emph{numeric drift}: percentages are occasionally re‑computed from rounded totals, yielding small but out‑of‑tolerance errors. This is most obvious in Stage B, where the model inserts neat-looking deltas that differ from the ground-truth bundle by 3–5\%.  
Second, \emph{fabrication under pressure}: Stage C asks for four numeric values per mode but offers no example. As a result, Gemini generates plausible-sounding values, yielding a perfect hallucination rate (HR = 1.00), despite structurally compliant prose.

Our experiments demonstrate that effective prompt design for numerical traffic summaries requires careful balance: bare structural prompts improve coverage but not fidelity; purely quantitative instructions without exemplars can overwhelm the model and backfire. In contrast, combining a concise, example-rich template with explicit numeric slots encourages Gemini 1.5 to replicate rather than fabricate figures—significantly improving numeric consistency and reducing hallucinations. However, these gains are realized only when stable metadata (e.g., camera-to-zone mappings) are established in advance and post-hoc validation is used to catch residual drift.

\subsubsection{Traffic Analysis Results}
\begin{figure}[!ht]
  \centering
  \includegraphics[width=0.9\textwidth]{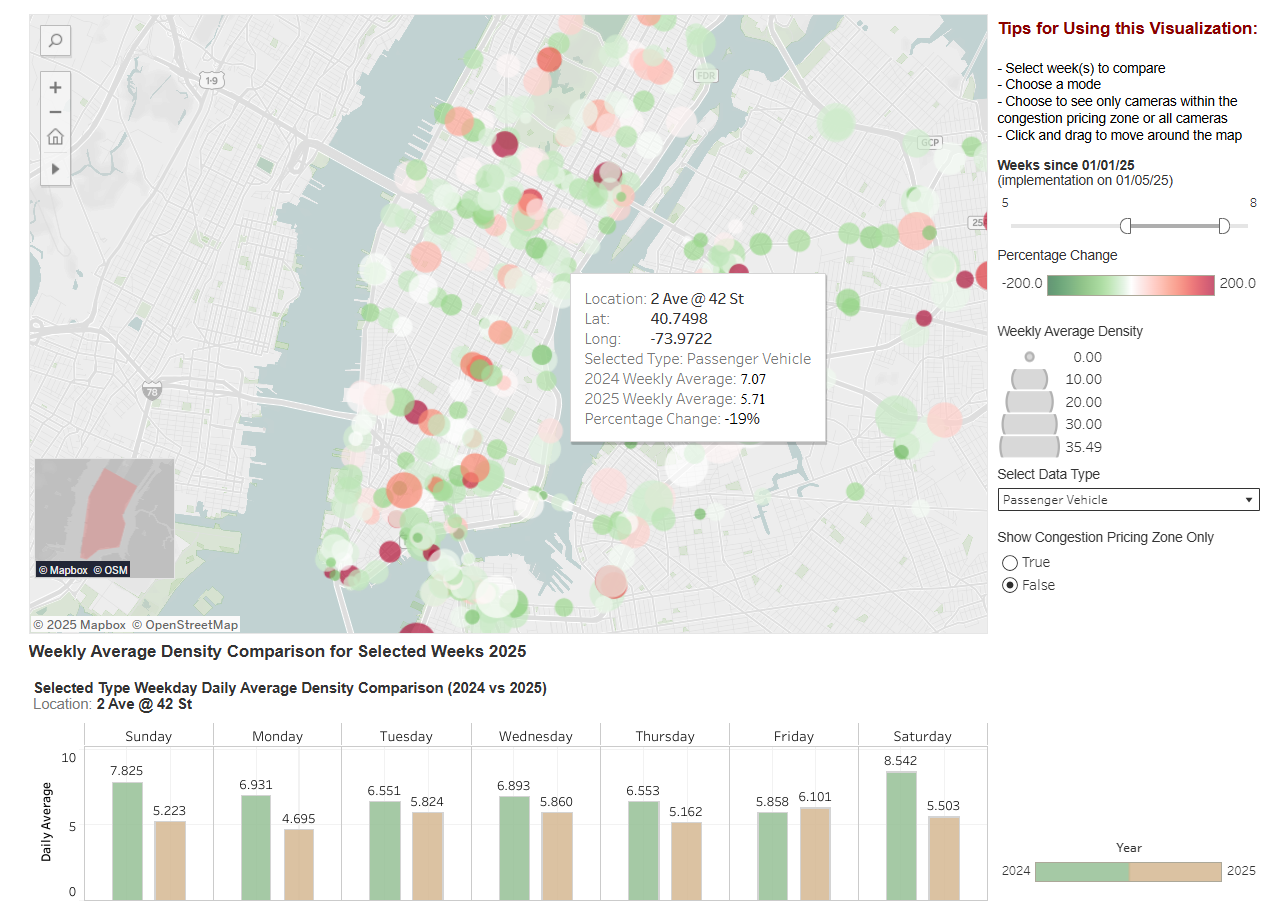}
  \caption{The multimodal density tracker integrates traffic camera feed data to support citywide and longitudinal traffic analysis}\label{fig:dashboard}
\end{figure}

The detection outputs were embedded into an interactive tool, the \textit{Multimodal Density Tracker}. \footnote{C2SMART Multimodal Density Tracker is available online: \url{https://c2smarter.engineering.nyu.edu/manhattan-congestion-tracker/}} The developed dashboard integrates camera geospatial metadata with weekly average traffic density metrics for four road user classes in 2024 and 2025 (Figure~\ref{fig:dashboard}). The interface enables users to visualize spatiotemporal variations by selecting specific weeks, view camera-level data on an interactive map, and access complementary bar charts comparing daily averages. Monthly summaries generated via LLM analysis provide contextual interpretation of changing traffic patterns by road user class. This user-friendly interface supports flexible exploration of both network-level trends and fine-grained local data, enabling a more intuitive and efficient interpretation of computer vision–derived traffic metrics. 

While spatial variation in \textbf{passenger vehicle} density was observed across locations, an overall average weekday change of -9\% was found within the CRZ during the first four months after the congestion pricing implementation. MTA data reported an average -11\% change in vehicles entering the CBD \cite{mta_cbdtp_2025}; While these metrics are not exactly the same, the smaller reduction in car density may point to a possible increase in activities within the CRZ. This is further supported by yellow taxi data, which showed an average 15\% rise in trips over the same period \cite{mta_cbdtp_2025}, with most yellow taxi activity concentrated in Manhattan’s CBD.

Prior to deployment, one hypothesis is that some drivers may adopt a park-and-ride strategy, opting to drive to areas near the CRZ boundary and switch to alternative modes such as transit. Our results showed a 2.7\% vehicle density increase in the buffer zone during the first month, indicating short-term diversion, but this effect faded, with the net change stabilizing at 0.1\% over the four months. These findings suggest that, while some initial traffic displacement to adjacent areas occurred, the overall impact on the buffer zone was limited in both magnitude and duration. 

\begin{figure}[!ht]
  \centering
  \includegraphics[width=1.0\textwidth]{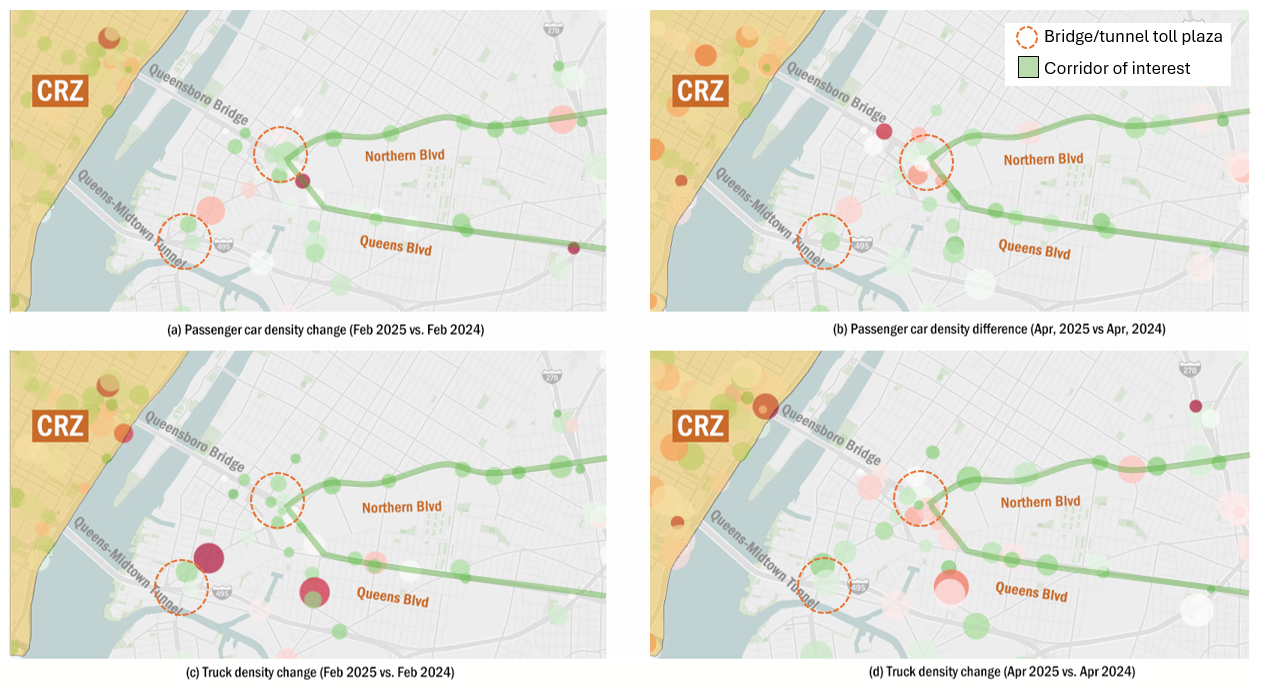}
  \caption{Example of corridor analysis}\label{fig:corridor}
\end{figure}

One key advantage of the dashboard is its ability to reveal directional trends in performance metrics over time while enabling spatial and temporal analyses at multiple scales (e.g., individual intersections, corridors, and broader zones; weekly and monthly intervals). For example, reductions in vehicle density in both passenger cars and trucks were detected along Queens Boulevard and Northern Boulevard, both key arterials linked to the access points connecting Queens and Manhattan, as indicated by green markers in Figure~\ref{fig:corridor}. These patterns suggest that the implementation of congestion pricing may have yielded non-CRZ corridor-wide benefits, reducing congestion not only at specific hotspots but across extended roadway segments. Furthermore, the observed reductions appear to persist across several consecutive months, indicating a degree of temporal stability in the intervention’s effects.

For \textbf{truck traffic}, substantial reductions in weekday truck density were observed within the CRZ, with declines of 16.6\% and 19.5\% during the first and second months following the deployment of congestion pricing, respectively. These effects need to be interpreted considering Manhattan’s unique geographic and infrastructural layout. Unlike many urban areas with multiple entry corridors, vehicular ingress to Manhattan is limited through a few crossings. Key access points within the CRZ include the Lincoln Tunnel, Holland Tunnel, Queens‑Midtown Tunnel, Queensboro Bridge, Manhattan Bridge, Brooklyn Bridge, Williamsburg Bridge, and Hugh L. Carey Tunnel, while the George Washington Bridge serves as a major connection point outside the tolled zone. Therefore, it is particularly important to examine whether truck traffic was rerouted to alternative crossings outside the toll zone. The results show that increases in truck density were detected in areas surrounding the George Washington Bridge and along the Cross Island Parkway near the Throgs Neck Bridge, suggesting potential redistribution or diversion of freight traffic to avoid the tolled area.

Another key finding is that while decreases in truck density continued in March and April (–6.3\%), the magnitude of reduction lessened. This suggests a potential rebound in truck activity within the CRZ. This trend may reflect operational adjustments by freight operators in response to the congestion pricing policy after 1-2  months of adaptation. Supporting this observation, a study by Geotab reported fewer large commercial trucks (GVWR Classes 3–8), with an increased presence of smaller multi-purpose vehicles, including last-mile delivery vans \cite{amny_geotab2025}. However, additional months of data are needed to determine whether this rebound is temporary or lasting.

For \textbf{pedestrians}, while February showed a slight decrease (-4.4\%) in average pedestrian density, overall we observed an average 4.0\% increase in weekday pedestrian density within the CRZ across January, March, and April, with increases of 4.1\%, 4.8\%, and 3.0\%, respectively. Weekend trends were more mixed. The highest pedestrian densities were observed near tourist/commercial locations and transit hubs, such as Times Square, SoHo, and Grand Central Terminal. While these changes may also be influenced by broader economic rebound \cite{nycedc_researchinsights}, the upward trend is notable, especially in January, which was significantly colder than normal, with average temperatures approximately 2.5 °F below the long-term historical average \cite{thestarryeye_january}.

A relatively consistent weekday increase of approximately 10\% in \textbf{cyclist density} was observed within the CRZ during the initial months of congestion pricing implementation. This increase is particularly notable in January, despite the colder and snowier weather conditions in 2025 compared to the same period in 2024. Similar to pedestrian activity, February was the only month that a reduction (-8.6\%) in cyclist density was found, which may be influenced by exogenous factors such as continued adverse weather conditions. 

In addition, combining traffic camera data with other available sources, such as link speed information \cite{manhattan_tracker}, can provide additional evidence of the impact of the program. For example, areas such as Midtown West (particularly along 8th, 9th, and 10th Avenues from 42nd Street to 57th Street), which experienced reductions in traffic density, also saw average speed improvements ranging from 1 to 7 mph. Crosstown streets also demonstrated moderate benefits from the program. However, avenues in Midtown and East Midtown exhibited more mixed outcomes following deployment, with certain locations, such as 5th Avenue, experiencing increased density and reduced travel speeds.

\subsubsection{Discussions and Limitations}
The proposed framework has the potential to contribute valuable insights to ongoing policy evaluation efforts by capturing traffic density patterns across the entire network, including within the CRZ. Our findings are generally consistent with trends reported by other agencies and studies \cite{mta_cbdtp_2025,amny_geotab2025, panynj_traffic_2025} and complement these existing data sources by offering more granular, location-specific information. In particular, the developed framework enables the observation of routing behaviors, such as shifts in truck traffic toward alternative crossings. Additionally, because conventional entry-based metrics do not capture trips that both originate and end within the zone, our approach helps fill a key gap by measuring \textbf{internal traffic activities} post-deployment. This added level of detail is particularly valuable given the limited availability of pre-deployment traffic volume and density data, providing a richer baseline and more continuous view of evolving travel patterns in response to congestion pricing.

While the developed framework demonstrates strong potential for large-scale traffic monitoring and data-driven policy evaluation, we have also noticed certain limitations. In particular, a few locations exhibited unusually high percentage changes in traffic density that may represent outliers or data anomalies rather than true behavioral shifts. Currently, these cases are either manually reviewed or retained as-is in the analysis, which may introduce noise in localized results. Incorporating automated anomaly detection methods into the pipeline would enhance the robustness of future analyses by systematically flagging such irregularities for further inspection. It is also important to note that the detection outputs should be interpreted with consideration of multiple external factors, such as weather, broader economic trends, seasonal fluctuations, and potential gradual adoption or rebound effects associated with the congestion pricing policy intervention.

\section{Conclusions and future Work}
This study presents a scalable, infrastructure-compatible framework for automated traffic monitoring and longitudinal performance evaluation using existing traffic camera networks. By integrating deep learning–based computer vision with a novel graph-based viewpoint normalization method, the system effectively addresses the dynamic viewpoint challenges posed by non-stationary PTZ cameras, enabling consistent and reliable traffic metric extraction over time. The object detection model achieved a mAP@0.5 at 0.788 for road user classes in a complex urban environment and with low image resolution. To further manage the complexity and volume of video-derived data, a multimodal LLM module with domain-specific prompt engineering was developed to produce interpretable, periodic summaries of traffic trends across multiple road user classes.

The operational applicability of the proposed framework was validated in a real-world deployment by analyzing 9 million images from approximately 1,000 traffic cameras in NYC during the initial months of its congestion pricing program. Traffic detection outputs were integrated into an interactive data visualization dashboard, the \textit{Multimodal Density Tracker}, which enables spatiotemporal analysis of road user density across multiple classes and time periods. The analysis revealed clear shifts in multimodal traffic patterns: a 9\% reduction in weekday passenger vehicle density within the CRZ, substantial decreases in truck density in early months (up to 19.5\%) with subsequent rebound patterns, and a consistent increase in cyclist and pedestrian activity, particularly in commercial and transit-oriented areas. An initial 2.7\% increase in adjacent buffer areas in the first post-deployment month was found, but stabilized over time. The flexibility of the framework allows for tracking traffic density at multiple geographic levels over time, providing insights into improved conditions on cross streets and in areas such as West Midtown, as well as benefits extending beyond the CRZ, with observed improvements along key arterials like Queens Boulevard and Northern Boulevard. Additionally, experiments on automatically generating monthly summaries using an LLM revealed that effective prompt design for numerical traffic pattern reporting requires a careful balance of structure and guidance, where example-rich templates with explicit numeric slots can improve fidelity and reduce hallucinations.

Collectively, these findings highlight the framework’s capacity to support high-resolution, timely policy evaluation at scale. By leveraging existing ITS infrastructure, the system offers a cost-effective and operationally feasible approach to traffic monitoring in dense urban settings. While this case study focuses on mobility and policy evaluation, the data generated by the proposed framework can also be extended to support other applications, such as traffic safety analysis and risk assessment. Future work will extend the system’s capabilities to support predictive analytics, automatic anomaly detection, and explore safety and behavioral modeling in response to diverse policy interventions.

\section{Acknowledgments}
The work in this paper is partially funded by C2SMART, a Tier 1 University Transportation Center at New York University. The contents of this paper only reflect the views of the authors who are responsible for the facts and do not represent any official views of any sponsoring agencies.%ChatGPT-4o was used only for language and grammar proofreading. The content, findings, and interpretations are entirely the authors' own and has been reviewed and approved by the authors prior to submission.

\section{Author Contributions}
The authors confirm contribution to the paper as follows: study conception and design: Jingqin Gao, Kaan Ozbay; data collection and experimental design: Donglin Zhou, Fan Zuo; analysis and interpretation of results: Jingqin Gao, Fan Zuo, Kaan Ozbay; draft manuscript preparation: Fan Zuo, Donglin Zhou, Jingqin Gao, Kaan Ozbay. All authors reviewed the results and approved the final version of the manuscript.

\bibliographystyle{unsrt}
\bibliography{references}
\end{document}